# Optimal Keywords Grouping in Sponsored Search Advertising under Uncertain Environments

Huiran Li and Yanwu Yang

School of Management, Huazhong University of Science and Technology

***Abstract:*** In sponsored search advertising, advertisers need to make a series of keyword decisions. Among them, how to group these keywords to form several adgroups within a campaign is a challenging task, due to the highly uncertain environment of search advertising. This paper proposes a stochastic programming model for keywords grouping, taking click-through rate and conversion rate as random variables, with consideration of budget constraints and advertisers' risk-tolerance. A branch-and-bound algorithm is developed to solve our model. Furthermore, we conduct computational experiments to evaluate the effectiveness of our model and solution, with two real-world datasets collected from reports and logs of search advertising campaigns. Experimental results illustrated that our keywords grouping approach outperforms five baselines, and it can approximately approach the optimum in a steady way. This research generates several interesting findings that illuminate critical managerial insights for advertisers in sponsored search advertising. First, keywords grouping does matter for advertisers, especially in the situation with a large number of keywords. Second, in keyword grouping decisions, the marginal profit does not necessarily show the marginal diminishing phenomenon as the budget increases. Such that, it's a worthy try for advertisers to increase their budget in keywords grouping decisions, in order to obtain additional profit. Third, the optimal keywords grouping solution is a result of multifaceted trade-off among various advertising factors. In particular, assigning more keywords into adgroups or having more budget won't certainly lead to higher profits. This suggests a warning for advertisers that it's not wise to take the number of keywords as the criterion for keywords grouping decisions.

# 1. Introduction

Sponsored search advertising has evolved into one of the most prominent online advertising channels [66]. Millions of advertisers choose search advertising to promote their products and services, taking advantage of precise targeting [9], low advertising costs [53] and high return on investment [28, 31]. Internet advertising revenues hit a record high of $107.5 billion in 2018, where search advertising accounts for 45.1% of that pie [25]. In sponsored search advertising, advertisers need to make a series of keyword related decisions. Indeed, keywords serve as a bridge linking advertisers, search users and search engines [65]. Different from other forms of online advertising, search advertisers have to organize keywords according to advertising structures defined by search engines. Well-organized keywords can secure more traffics and revenues through serving the right ads to the right customers [63]. Therefore, for search advertisers, how to effectively organize keywords of interest in their campaigns is a critical issue.

Throughout the entire lifecycle of search advertising campaigns, advertisers have to face a series of keyword related decisions, namely keyword generation, selection, grouping and adjustment [65]. Current research efforts along the line of keyword research mainly focus on keyword generation (e.g., [42, 43, 68]) and keyword selection (e.g., [29, 33]). From the operational perspective, there is yet to explore keyword decisions on how to organize keywords according to advertising structures defined by search engines.

Sponsored search advertising campaign development involves organizing keywords into adgroups, developing adcopies for adgroups [8]. In the search advertising structure employed by major search engines (e.g., Google, Bing), under an advertiser's account, one or several campaigns run simultaneously to fulfill a certain promotional goal, where each campaign includes one or several adgroups, and each adgroup in turn contains one or more adcopies and a shared set of keywords. Naturally, adgroup is prevalent as the basic unit for daily advertising operations. First, Bing Ads (2019) [5] states that adgroups are the best way to organize advertising campaigns. In particular, adgroup allows advertisers to better track the effectiveness of their advertising efforts [19]. Second, an advertiser needs to build a keyword list for each adgroup from a predefined set of keywords of interest in order to precisely display her ads to the targeted consumers [20]. When a searcher's query matches one or more keywords in an adgroup, its associated advertisement will be triggered to appear on the search engine results pages. Thus, organizing keywords with adgroups allows advertisers to show their advertisements to customers searching for similar things. Contrarily, for an advertiser, if her adgroups aren't set up right, she'll hardly reach target population with her advertising campaigns. Third, instead of optimizing or predicting the performance for each individual keyword, it's more convenient for advertisers to manage several adgroups [24]. Moreover, by aggregating keywords into groups, advertisers can refine strategies of allocating advertising spendings over long-tail keywords [47]. Last but not least, the result of keywords grouping decisions provides a basic foundation and sense of organization as advertisers move into the ad-writing phase of their campaigns. That is, well-organized keywords make advertisers easier to compose adcopies and provide what consumers needed during a search session [7], which ensures more relevant user experiences [35]. Therefore, keywords grouping is just as important as any of those other keyword decisions in sponsored search advertising [65, 21]. Hereafter, in this paper, we use "keywords grouping" to refer to the decision that organize keywords at the adgroup level within an advertising campaign.

Most of current keywords grouping practices operate on an aggregate level [46]. For example, adgroups are usually organized based on product (or service) lines that advertisers want to promote [18]. As reported in [2], advertisers might also categorize their keywords according to keyword characteristics (i.e., generic, branded or retailered) while creating adgroups. Such approaches typically consume a lot of manual efforts, and hardly obtain an optimal solution for keywords grouping because they are contingent on semantic relationships between keywords and advertising themes, which are largely prone to subjective errors. We argue that a scientific solution for keywords grouping should be able to find an optimized strategy to segment a set of keywords into groups to fill a certain promotional goal (e.g., to maximize the expected profit), with consideration of related advertising factors (e.g., CPC and CTR) and budget constraints. From an operational perspective, keywords grouping is an overlooked part in the search advertising workflow [57]. In search advertising campaign management, the AB test is an experimental approach, through which advertisers isolate different factors to see which adcopy performs best [56]. However, it's an impossible task to apply the AB test to figure out the optimal keywords grouping solution because the computational space exponentially increases with the number of keywords. Although there is an urgent and practical need for keywords grouping, to the best of our knowledge, the literature in this direction is sparse. This paper aims to bridge this critical gap.

There are many challenges associated with keywords grouping decisions. On one hand, the search advertising environment is highly uncertain [45, 60]. That is, advertisers have to make keywords grouping decisions before obtaining values of keyword performance indexes. As [36] stated, IT-based high-technology industries share common characteristics, which are featured by market uncertainty, technology uncertainty, and competitive volatility [59]. Likewise, search advertising also suffers from three types of uncertainties: disturbance coming from market noise (i.e., social hot news makes search volume and click amount of some keywords increase sharply), uncertainty stemming from technology evolutions (i.e., search engines improve their ranking algorithms for advertising display), and uncertainty originating from competitive volatility (i.e., advertisers can adjust their strategy arbitrarily). On the other hand, advertisers, especially those from small enterprises, usually face serious budget constraints [61, 62]. It implies that advertisers need to appropriately group keywords under a limited budget in order to maximize their advertising performance. The uncertainties of search advertising are reflected by several performance indexes. More specifically, click-through rate (CTR) and conversion rate (CVR) vary drastically over keywords and are often unknown in advance, which raises large uncertainties for keywords grouping decisions [16]. This motivates us to study the keywords grouping problem in a stochastic model. Our premise is that advertisers can obtain limited amount of information about the range of values taken by factors of interest (e.g., probability distributions) by analyzing historical reports from search advertising campaigns. The intended work is different from the above keywords grouping approaches in twofold. First, our approach considers uncertainties in search advertising markets and conducts risk management for advertisers with different risk preferences. Second, our approach based on the branch and bound algorithm can obtain the optimal solution by traversing the solution space of keywords grouping decisions.

In this work, we formulate a stochastic model for keywords grouping to maximize the expected profit from a search advertising campaign. In particular, our model takes CTR and CVR as random variables[1]. First, we use the concept of chance constraint to describe the probability of meeting the budget constraint within a certain degree. Second, the variance of profit over a unit of budget at the campaign level is used to measure an advertiser's risk, in order to balance expected profit and risk exposures. Then we develop a branch-and-bound solution process to solve our keywords grouping model. Furthermore, we conduct computational experiments to evaluate the performance of our keywords grouping model with two real-world datasets collected from field reports and logs of search advertising campaigns, by comparing with five baselines. Among them, the first two are commonly used in practice, and the third and the fourth are derived from the extant literature, and the fifth is a deterministic approach derived from our approach. The first baseline represents the case that the advertiser puts all keywords into a single adgroup (i.e., BASE1-Nogrouping). The second baseline subdivides keywords according to the advertiser's products to be promoted to form adgroups (i.e., BASE2-Product). The third applies k-means clustering to segment keywords (i.e., BASE3-Kcluster). The fourth categorizes keywords according to a keyword hierarchy based on semantic relationships (i.e., BASE4-Hierarchy). The fifth baseline assigns keywords into adgroups according to their profits in a greedy manner, by following a deterministic model derived from our stochastic keywords grouping model developed in Section 4 (i.e., BASE5-Profit).

---

[1] Note that CTR and CVR might be, more or less, improved through keywords grouping strategies, however, which is not the ultimate goal, but intermediary performance indexes, for advertisers. Moreover, neither CTR nor CVR provides comprehensive clue for the ultimate advertising goal [26]. In this work, we distinguish the CTR (CVR) inherited in keywords themselves and the CTR (CVR) raised by adgroups, and use the product of them to represent the CTR (CVR) of a keyword assigned in an adgroup.

Experimental results show that a) our keywords grouping approach outperforms five baselines in terms of the profit and ROI, with relatively lower risks; b) compared to five baselines, our approach assigns more keywords and can approximately approach the optimum in a steady way; c) in the keywords grouping decisions, as the budget increases, the profit grows accordingly; however, the marginal profit does not necessarily show the marginal diminishing effect, i.e., it does not always decrease with the increase of the budget; d) assigning more keywords into adgroups won't certainly lead to a higher profit. Essentially, the optimal keywords grouping solution is a result of multifaceted trade-off among various advertising factors.

These findings provide critical managerial insights for advertisers in sponsored search. First, keywords grouping is a critical advertising decision that cannot be overlooked, especially under this more complicated market environment with a large number of keywords. Second, increasing the budget in keywords grouping decisions can be a worthy try for advertisers to obtain additional profit. Third, this research suggests a warning for advertisers that it's not wise to take the number of keywords as the criterion for keywords grouping decisions.

The key contributions of our study include the following. From the academic perspective, to our knowledge, this is the first study on keyword grouping decisions. In the extant literature, few researches were reported to deal with keywords decisions following the search advertising structure that is differ from flat structure of traditional advertising. Our study explores the keyword grouping problem in sponsored search advertising, with consideration of the advertising structures defined by major search engines. Moreover, the outcome of our research sheds lights on keyword research in online advertising. This study shows that keyword grouping also matters for advertisers in sponsored search advertising. And it matters in some ways that have not been previously recognized in the extant literature. We also find that the marginal profit does not necessarily show the marginal diminishing effect with the increase of the budget. These findings facilitate a better understanding on keywords research and search advertising structures. From a substantive point of view, our findings offer concrete managerial insights for advertisers to make practical keyword decisions in sponsored search advertising. From the methodology perspective, we develop a stochastic model for keyword grouping decisions and correspondingly develop a branch and bound algorithm to solve our model. Moreover, our keywords grouping model and the corresponding branch-and-bound solution can be generalized to segmentation problems in keywords research and other similar decision scenarios of keyword-related advertising forms (e.g., Twitter Ads).

The reminder of this paper is structured as follows. First, we provide a brief literature review in Section 2. In Section 3, we build a stochastic keywords grouping model in search sponsored advertising. Section 4 presents a branch-and-bound algorithm to solve our model. We report experimental results and discuss related managerial implications in Section 5. Finally, we conclude this research in Section 6.

## 2. Related Work

In this section, we first focus on search advertising decisions, then narrow down to keyword related decisions, with special attention to keywords grouping decisions in the context of sponsored search advertising.

### 2.1. Search Advertising Decisions

Most current research efforts in the field of search engine advertising focus on three categories: search auction mechanism design and equilibrium analysis, empirical studies on different factors, and strategic advertising behaviors and decisions [64]. It is of great important significance for advertisers to develop optimal advertising strategies facilitating campaign manipulations [49]. A majority of research on search advertising decisions has focused on bidding strategy [4, 10, 14, 50, 52, 67], budget allocation [62], and keyword decisions [1, 38, 44, 47, 65].

From the perspective of search engines, theoretical and empirical analyses in [15] suggested that strategic behaviors are widespread and costly, and a switch to a VCG-based mechanism might stabilize auction outcomes with neutral or even positive effects on revenues, at least relative to the old Overture mechanism that was based on the first-price auction. By considering bid dynamics and rankings of advertisers, [69] proposed a dynamic model and identified an equilibrium bidding strategy. Their empirical framework, based on a Markov switching regression model, suggested the existence of cyclical bidding strategies. Using a game-theoretic model, [4] examined the strategic role of keyword management costs aroused from advertisers' decisions (e.g., the keywords they choose to bid on and their bidding prices) and of broad match which automates bidding on keywords, in sponsored search advertising. Their analysis showed that the search engine will increase broad match bid accuracy up to the point where advertisers choose broad match, but increasing the accuracy any further reduces the search engine's profits. Through building a Hierarchical Bayesian model to address the endogeneity problem and using the Markov Chain Monte Carlo (MCMC) method to identify the parameters, [14] empirically explored how to manage ad campaigns when advertisers have to bid on multiple keywords. The results suggested that it is important to differentiate among the various bidding strategies for various keyword categories and match types. [69] modeled the budget-constrained bidding as a stochastic multiple-choice knapsack problem, and designed an algorithm that selects items online based on a threshold function which can be built based on historical data. Their algorithm achieved about 99% performance compared to the offline optimum when applied to a real bidding dataset. Another stream of search advertising decision research is budget allocation. Effectively allocating the limited advertising budget is a critical search advertising decision. With multiple search advertising markets and a finite time horizon, [62] developed a novel budget allocation optimization model. A customized advertising response function was proposed when considering distinctive features of sponsored search, including the quality score and the dynamic advertising effort. [37] explored how to distribute advertising budget over the keywords of their interest in order to maximize their return. The results showed that simple prefix strategies that invest on all cheap keywords up to some levels are either optimal or good approximations for many cases.

However, as a matter of fact, advertisers are not allowed to spread their budget across keywords directly in actual sponsored search advertising.

In the next section, we narrow down to keyword related decisions in sponsored search advertising.

## 2.2. Keyword Decisions

Keywords serve as an essential bridge linking advertisers, search users and search engines in sponsored search advertising. Advertisers have to deal with a series of keyword decisions throughout the entire lifecycle of search advertising campaigns, including keywords generation, selection, grouping and adjustment [65]. They developed an integrated multi-level computational framework for keyword optimization (MKOF) supporting a set of strategies across different levels of abstractions (e.g., domain, market, campaign, adgroup and keyword) throughout the lifecycle of sponsored search advertising campaigns. Moreover, advertisers have to monitor the realtime performance of advertising campaigns and adjust their keyword decisions accordingly. Existing research on keyword related decisions primarily focuses on the first two issues.

Keywords generation can be categorized into three streams, i.e., query log-based, proximity-based and meta-tag crawlers-based methods [1, 42]. In the branch of query log-based methods, keywords are mainly suggested by conducting association/co-occurrence analysis in search engine query logs [34, 68, 70]. Proximity-based keyword generation methods query search engines with the seed keyword and recommend keywords from the query results possessing high proximity to the seed keyword [1, 58]. In addition, some efforts calculate the proximity based on vocabulary dictionaries/corpus pre-constructed by domain experts [11], e.g., thesaurus dictionary, Wikipedia, etc. The meta-tag crawlers based methods focus on finding relevant keywords from meta-tags. They send the seed keyword to the search engine and extract meta-tag keywords from the top ranked web pages [42]. Some popular online tools like WordStream and Wordtracker use meta-tag crawlers to search meta-tag keywords and make suggestions of relevant keywords for advertisers.

Selecting the most appropriate keywords after keyword generation help prevent advertisers from targeting wrong groups of consumers and eventually wasting their advertising budget with poor returns [27]. [45] selected keywords by ranking them on their profit-to-cost ratio which guarantees the conversion of the average expected profit to a near-optimal solution. [29] proposed to optimize advertising keywords with feature selection techniques applied to the set of all possible word combinations comprising past users' queries. [17] proposed an automatic way of examining keyword ambiguity based on probabilistic topic models from machine learning and computational linguistics. The study can also help search engines design keyword planning tools to aid advertisers when choosing potential keywords. Keyword characteristics and related performance prediction are a prerequisite for keyword selections. [14] used a hierarchical Bayesian estimation framework to identify a system of simultaneous equations and derive results related to how keyword categories and keyword match types are associated with performance outcomes (e.g., such as CTR, CVR, CPC, and rank). [22] presented a relevance model that could accurately predict relevance for query-ad pairs, by incorporating implicit relevance feedback from sparse user clicks in search logs.

After a set of keywords of interest is obtained, advertisers need to group these keywords according to advertising structures to form adgroups. However, few prior studies have been carried out on this topic. A related research area to keywords grouping is keyword clustering. [44] used a hierarchical clustering of terms based on the partitioning of a keyword-advertiser matrix as a source of term relationships to predict click-through rate in sponsored search advertising. [39] applied a k-means clustering algorithm to categorize the referral keywords with similar characteristics of onsite customer behavior (e.g., click-through rate and revenue). They demonstrated how online businesses can leverage this segmentation clustering approach to provide a more tailored consumer experience and develop better business models to increase advertising conversion rates.

## 2.3. Stochastic Optimization in Search Advertising

Stochastic optimization models have been developed to deal with uncertainties in search advertising decisions. [37] formulated a stochastic optimization model to distribute the budget over keywords, based on natural probabilistic models of distribution over future queries. [40] developed an analytical auctions model centered on an advertiser facing multiple unknown opponents who submit random bids. Assuming that the probability of click comes from independent stochastic processes, they alluded that the actual CTR and its estimated weight is affected by the type of searchers. [12] considered the ad position as a random function influenced by bidding prices, and constructed a stochastic model for allocating the daily budget over a set of keywords, which use the beta density to represent the random ad position given a fixed bidding price.

In sponsored search advertising, keywords grouping decisions are influenced by many factors (e.g., CTR, CVR) that cannot be known precisely in advance. This motivates us to explore the keywords grouping problem in the stochastic setting with consideration of budget constraints of adgroups and advertisers' risk-tolerances. In this work, we are intended to explore the problem of keywords grouping under uncertainty environment. To the best of our knowledge, this is the first research effort in this direction.

## 3. The Model

In this section, we build a stochastic model for keywords grouping to maximize the expected profit in sponsored search advertising, with consideration of budget constraints of adgroups and advertisers' risk-tolerances. There might be other constraints for advertisers (e.g., geography and time), our research considers budget constraints and risk constraints that are commonly taken into account in prior work (e.g., [51, 60]). The keyword decision scenario under consideration by this research is: for an advertiser, given a set of campaign-specific keywords, how to group these keywords into several adgroups. The notations used in this paper are listed in Table 1.

### 3.1 The Objective

Let $d_i$ denote the total number of search demands of the $i^{th}$ keyword in a search market. The search demand of a keyword is defined as the total number of queries triggered from it. Let $c_{ij}$ denote the click-through rate (CTR) of the $i^{th}$ keyword in the $j^{th}$ adgroup. Given an advertising campaign with $m$ adgroups and a set of keywords (i.e., $n$), the decision variable $x_{ij}$, $j = 1, \ldots, m$, $i = 1, \ldots, n$, indicates whether the $i^{th}$ keyword is assigned to the $j^{th}$ adgroup or not, i.e.,

$$x_{ij} = \begin{cases} 1, & \text{if the } i^{th} \text{ keyword is assigned to the } j^{th} \text{ adgroup}; \\ 0, & \text{otherwise}. \end{cases}$$

Let $p_{ij}$ denote the cost-per-click (CPC) of the $i^{th}$ keyword in the $j^{th}$ adgroup. According to major search advertising structures, advertisers can set the max CPC on both the adgroup and keyword levels. In our research, for the keywords grouping problem, we use $p_{ij}$ on the keyword level. Then the cost for a campaign is $\sum_{j=1}^{m}\sum_{i=1}^{n} x_{ij} d_i c_{ij} p_{ij}$. Let $r_{ij}$ and $v_i$ denote the conversion rate (CVR) and value-per-sale, respectively. Thus, the profit of an advertising campaign can be represented as $z(x_{ij}) = \sum_{j=1}^{m}\sum_{i=1}^{n} x_{ij} d_i c_{ij} (r_{ij} v_i - p_{ij})$. In this research, we use CTR $c_{ij}$ and CVR $r_{ij}$ as random vectors to capture uncertainties in searchers' behaviors, advertising market volatility, etc. So $z(x_{ij})$ is also a random variable. Therefore, the objective of keywords grouping decisions is to maximize the profit expected in an ad campaign, given by $E[z(x_{ij})] = E[\sum_{j=1}^{m} \sum_{i=1}^{n} x_{ij} d_i c_{ij} (r_{ij} v_i - p_{ij})]$.

### 3.2 The Budget Constraint

Advertisers usually have a limited budget for search advertising. We can naturally assume that the budget is less than a sufficient amount. Let $B_j > 0$ denote the advertising budget available to a given adgroup $j$, then we have $\sum_{i=1}^{n} x_{ij} d_i c_{ij} p_{ij} \leq B_j$.

Due to the stochastic nature of $c_{ij}$, the budget constraint can be represented as a chance constraint, i.e., $P\{\sum_{i=1}^{n} x_{ij} d_i c_{ij} p_{ij} \leq B_j\} \geq \alpha_j$, where the probability that the cost of adgroup $j$ is less than the allocated budget, is greater than or equal to a certain level $\alpha_j$ (i.e., an acceptable probability range). In order to simplify the expression, we also treat the cost of a keyword $s_i = \sum_{j=1}^{m} d_i c_{ij} p_{ij} x_{ij}$ as a random variable. Then we have $P\{\sum_{i=1}^{n} x_{ij} s_i \leq B_j\} \geq \alpha_j$.

### 3.3 The Risk Constraint

Our keywords grouping model also considers different risk preferences from advertisers. A risk-averse advertiser prefers certainty to risk, and low risk to high risk, thus prefers a strategy within her risk tolerance; while a risk-loving advertiser would prefer the chance of getting more revenue at the cost of high risk; A risk neutral advertiser would not have any preference. [23] stated that the profit variance can be interpreted as a risk measure in advertising market. Following [60], in order to balance the expected profit and risk exposures, we take the variance of profit $z(x_{ij})$ over a unit of budget as the risk, given as

$$\frac{Var(z(x_{ij}))}{\sum_{j=1}^{m} B_j} = \frac{Var(\sum_{j=1}^{m} \sum_{i=1}^{n} x_{ij} d_i c_{ij} (r_{ij} v_i - p_{ij}))}{\sum_{j=1}^{m} B_j} \leq \theta$$

where $\theta$ is the risk-tolerance of an advertiser.

### 3.4 The Stochastic Keywords grouping Model

In summary, the keywords grouping problem can be formulated as the following stochastic model:

$$\max \quad E\left[\sum_{j=1}^{m} \sum_{i=1}^{n} x_{ij} d_i c_{ij} (r_{ij} v_i - p_{ij})\right]$$

$$\text{s.t.} \quad P\left\{\sum_{i=1}^{n} x_{ij}d_i c_{ij} p_{ij} \leq B_j\right\} \geq \alpha_j$$

$$\text{Var}\left(\sum_{j=1}^{m}\sum_{i=1}^{n} x_{ij}d_i c_{ij}(r_{ij}v_i - p_{ij})\right) / \sum_{j=1}^{m} B_j \leq \theta \quad (1)$$

$$\sum_{j=1}^{m} x_{ij} \leq 1$$

$$x_{ij} = 0/1, d_i \geq 0, v_i \geq 0, p_{ij} \geq 0$$

$$i = 1, \ldots, n, j = 1, \ldots, m$$

Model (1) is a stochastic 0-1 multiple knapsack programming problem where the decision variable $x_{ij}$ is binary. In our keywords grouping model, the objective function aims to maximize the expected profit from all adgroups. The first constraint is a series of chance constraints of soft budgets for adgroups, the second constraint describes the risk-tolerance constraint, and the third constraint requires that a keyword can only belong to no more than one adgroup. In search advertising, advertisers might assign several identical keywords in two or more adgroups. However, in the case that multiple keywords match a same search term, search engines have a set of preferences determining which keyword is used to trigger an ad into an auction. As such, it's not a good option to have identical keywords in different groups, because advertisers may lose track over which of their keywords works [55].

## 4. Solution

In this section, we provide a branch-and-bound algorithm solution process to solve our keywords grouping model (1). Branch and bound is an enumerative method that has been used successfully to solve a variety of combinatorial problems by reducing search and computational efforts necessary to find the optimum [13, 30]. In particular, given a certain fixed budget for each adgroup in a campaign and the advertiser's risk-tolerance, an optimal solution should adaptively assign a set of keywords to several adgroups to maximize the expected profit. In order to simplify the computation, we reduce the dimensionality of the decision variable by taking the keyword-adgroup combination as the basic unit for decision variables. Then we have $n * m$ keyword-adgroup combinations. In the following we present details of our branch-and-bound algorithm solution process for our keywords grouping model. For more details on branch-and-bound algorithm, refer to see [30].

First, we use a stochastic simulation to check whether the chance constraint of budget is satisfied for each adgroup, which is given in Algorithm SSCCAB (standing for stochastic simulation for chance constraints of advertising budget). When assigning a keyword into adgroup $j$, if and only if (iff) the total cost is less than the budget constraint within confidence interval for adgroup $j$, i.e., $P\{\sum_{i=1}^{n} x_{ij} s_i \leq B_j\} \geq \alpha_j$, then the indicative variable $\hat{x}_{ij} = 1$, otherwise 0.

| **Algorithm (SSCCAB)** |
|---|
| **Input:** |
| $\{i \mid i = 1,2, \ldots, n\}$ – a set of keywords of interest |
| $\{j \mid j = 1,2, \ldots, m\}$ – a set of adgroups |
| $B_j$ – the soft budget constraint for the $j^{th}$ adgroup |
| $\alpha_j$ – the prescribed probability of budget constraints for the $j^{th}$ adgroup |
| $s_i \sim N(\mu_i, \sigma_i^2)$ – the distributions of the $i^{th}$ keyword cost |
| **Output:** $\hat{x}_{ij}$ – the binary variables indicated whether the chance constraints of budget are satisfied when assigning the $i^{th}$ keyword into the $j^{th}$ adgroup |
| **Procedure:** |
| 1. Let $t' = 0$. |
| 2. Extract a set of values for the cost of keywords $s' = \{s'_1, s'_2, \ldots, s'_n\}$ from the corresponding distribution of $s_i \sim N(\mu_i, \sigma_i^2)$ as a sample. |
| 3. **If** $\sum_{i=1}^{n} x_{ij} s'_i \leq B_j$ **then** we have $t'++$. |
| 4. **Repeat** steps 2 and 3 for $t$ times. |
| 5. $\alpha'_j = t'/t$. |
| 6. **If** $\alpha'_j \geq \alpha_j$ **then** $\hat{x}_{ij}=1$; **else** $\hat{x}_{ij}=0$. |

Next, we calculate the upper bound for the branch and bound algorithm through continuous relaxation of model (1). Specifically, we relax $x_{ij}$ from a binary variable in $\{0,1\}$ to a continuous variable in $[0,1]$. Following [41], it is known that the set defined by constraint $P\{\sum_{i=1}^{n} x_{ij} s_i \leq B_j\} \geq \alpha_j$ is convex if function $\sum_{i=1}^{n} x_{ij} s_i$ is quasi-convex and $s_i$ has a log-concave density. The first property can easily be proved as our function $\sum_{i=1}^{n} x_{ij}s_i$ is linear, thus it is quasi-convex. With regard to the second property, according to [12], the number of clicks per impression $c_{ij}$ (i.e., CTR), has dimensions [click/impr]. It is a Bernoulli random variable with parameter $p(x_{ij})$ representing the possibility of an advertisement associated with the $i^{th}$ keyword being clicked if it is assigned to the $j^{th}$ adgroup. Then the number of clicks of the $i^{th}$ keyword $C_i$ is a binomial random variable with parameters $(d_i, p(x_{ij}))$. The binomial can be accurately approximated by the normal provided that $d_i \cdot p(x_{ij}) \geq 10$ and $d_i \cdot [1 - p(x_{ij})] \geq 10$. Such that we naturally assume that the random variable $C_i$ (i.e., the number of clicks) is normal[2]. Thus, the cost $s_i$ of keyword $i$, i.e., the product of the number of clicks (the random variable) and the average cost per click (constant), is also independently normally distributed. The second property can be proved for normal distributions. This means that the chance constraint $P\{\sum_{i=1}^{n} x_{ij}s_i \leq B_j\} \geq \alpha_j$ defines a convex set in the special case of a relaxed keywords grouping problem with normally distributed costs.

Then we can solve the continuous chance-constraint keywords grouping model by reformulating it as an equivalent, deterministic second-order-cone-programming (SOCP) problem [32]. From search advertising logs and reports, we can get the mean $\mu_i$ and standard deviation $\sigma_i$ of $s_i$. As $B_j$ is a constant, $Var[B_j] = 0$, $E[B_j] = B_j$. Then we have

$$\frac{\sum_{i=1}^{n} x_{ij}s_i - B_j - (\sum_{i=1}^{n} x_{ij}E[s_i] - B_j)}{\sqrt{\sum_{i=1}^{n} x_{ij}^2 Var[s_i]}},$$

which represents a standard normal variant.

The inequality $\sum_{i=1}^{n} x_{ij}s_i \leq B_j$ is equivalent to

$$\frac{\sum_{i=1}^{n} x_{ij}s_i - B_j - (\sum_{i=1}^{n} x_{ij}E[s_i] - B_j)}{\sqrt{\sum_{i=1}^{n} x_{ij}^2 Var[s_i]}} \leq -\frac{\sum_{i=1}^{n} x_{ij}E[s_i] - B_j}{\sqrt{\sum_{i=1}^{n} x_{ij}^2 Var[s_i]}}.$$

Then the chance constraint $P\{\sum_{i=1}^{n} x_{ij}s_i \leq B_j\} \geq \alpha_j$ is equivalent to

$$P\left\{\eta \leq -\frac{\sum_{i=1}^{n} x_{ij}E[s_i] - B_j}{\sqrt{\sum_{i=1}^{n} x_{ij}^2 Var[s_i]}}\right\} \geq \alpha_j,$$

where $\eta$ obeys a standard normal distribution.

---

[2] We implicitly assume that the parameter $d_i$ is reasonably large so that the two conditions given above are satisfied.

Therefore, chance constraints are established, if and only if the following condition is satisfied.

$$\phi^{-1}(\alpha_j) \leq -\frac{\sum_{i=1}^n x_{ij} E[s_i] - B_j}{\sqrt{\sum_{i=1}^n x_{ij}^2 Var[s_i]}}$$

$$\Rightarrow \sum_{i=1}^n x_{ij} E[s_i] + \phi^{-1}(\alpha_j) \sqrt{\sum_{i=1}^n x_{ij}^2 Var[s_i]} \leq B_j$$

$$\Rightarrow \sum_{i=1}^n x_{ij} \mu_i + \phi^{-1}(\alpha_j) \sqrt{\sum_{i=1}^n x_{ij}^2 \sigma_i^2} \leq B_j$$

We can construct a concave optimization model through relaxing $x_{ij}$ to a continuous variable in [0,1] and transforming the chance constraints of budget into a deterministic formulation. It is given as,

$$\max \quad E\left[\sum_{j=1}^m \sum_{i=1}^n x_{ij} d_i c_{ij} (r_{ij} v_i - p_{ij})\right]$$

$$\text{s.t.} \quad \sum_{i=1}^n x_{ij} \mu_i + \phi^{-1}(\alpha_j) \sqrt{\sum_{i=1}^n x_{ij}^2 \sigma_i^2} \leq B_j \qquad (2)$$

$$Var\left(\sum_{j=1}^m \sum_{i=1}^n x_{ij} d_i c_{ij} (r_{ij} v_i - p_{ij})\right) / \sum_{j=1}^m B_j \leq \theta$$

$$\sum_{j=1}^m x_{ij} \leq 1$$

$$0 \leq x_{ij} \leq 1, d_i \geq 0, v_i \geq 0, p_{ij} \geq 0$$

$$i = 1, \ldots, n, j = 1, \ldots, m$$

In the following, we employ the interior point method [54] to find the optimal solution for model (2). From this, we can get the upper bound, i.e., SUP. The overall framework of branch and bound algorithm to solve our keywords grouping model is given in Algorithm BBKG (standing for the branch and bound algorithm for keywords grouping).

| **Algorithm (BBKG)** |
|---|
| **Input:** |
| $\{i \mid i = 1, 2, \dots, n\}$ – a set of keywords of interest |
| $\{j \mid j = 1, 2, \dots, m\}$ – a set of adgroups |
| $B_j$ – the soft budget constraint for the $j^{th}$ adgroup |
| $d_i$ – the search demand of the $i^{th}$ keyword |
| $c_{ij}$ – the click-through rate (CTR) of the $i^{th}$ keyword in the $j^{th}$ adgroup |
| $r_{ij}$ – the conversion rate (CVR) of the $i^{th}$ keyword in the $j^{th}$ adgroup |
| $v_i$ – the value-per-sale (VPC) of the $i^{th}$ keyword |
| $p_{ij}$ – the cost-per-click (CPC) of the $i^{th}$ keyword in the $j^{th}$ adgroup |
| $\theta$ – the risk-tolerance |
| **Output:** $x_{ij}$ – the decision variable indicated whether the $i^{th}$ keyword is assigned to the $j^{th}$ adgroup. |
| **Procedure:** |
| 1) Sort adgroups according to decreasing $B_j$, sort keywords according to decreasing $E[d_i c_{ij}(r_{ij} v_i - p_{ij})]$, and Keywords_Grouping_List = $\emptyset$. |
| 2) **For** adgroup $j$ from 1 to m |
|    **for** keyword $i$ from 1 to n |
|    **if** $\hat{x}_{ij} = 1$, $\text{Var}(\sum_{j=1}^{m} \sum_{i=1}^{n} x_{ij} d_i c_{ij}(r_{ij} v_i - p_{ij}))/\sum_{j=1}^{m} B_j \leq \theta$ and $\sum_{j=1}^{m} x_{ij} \leq 1$ then $x_{ij} = 1$, INF = max {the expected profit}, add the feasible solution to Keywords_Grouping_List, and the upper bound SUP = $\infty$. |
| End for |
| End for |
| 3) **If** Keywords_Grouping_List = $\emptyset$ then go to step 7, **else** current_solution = solution in Keywords_Grouping_List with maximum expected profit, go to step 4. |
| 4) **If** SUP > INF for current_solution then go to step 5, **else** delete the solution from the list then go to step 3. |
| 5) **If** there is no accepted keyword left in the selected solution that does not already have a plunged or rejected subset, then delete the solution from the list then go back to step 3, **else** following the ranking, choose the first accepted keyword that does not already have |

> a plunged or rejected subset calculate SUP for the subset defined by rejecting this keyword, go to step 6.
>
> 6) **If** SUP ≤ INF then delete this subset go to step 5; **else** plunge the subtree as described in 2 and add the found branch together with the value SUP to the Keywords_Grouping_List.
>
> **If** the expected profit of this solution > INF, then update INF, go to step 3.
>
> 7) Return the corresponding keywords grouping result $x$.

Algorithm BBKG searches the complete space of solutions for the optimal keywords grouping solution within budget chance-constraints and risk-tolerance. The keywords grouping solution is a n * m 0-1 matrix. At any point during the process, the status with respect to the search of the keywords grouping solution space is described by a pool of yet unexplored subsets of the space and the best keywords grouping solution found so far. Initially, only one subset exists, namely the complete solution space, and the best solution found so far is ∞. The unexplored subspaces are represented as nodes in a dynamically generated search tree, which initially only contains the root, and each iteration of a keywords grouping branch and bound algorithm processes one such node. The iteration has three main components: selection of the node to process, bound calculation, and branching. Our strategy for selecting the node to process is in descending order of expected keyword profit. The operation of an iteration after choosing the node is branching, i.e. subdivision of the solution space of the node into $m + 1$ subspaces (i.e., $m + 1$ represents the cases that the keyword is assigned into one of the m adgroups or no adgroup) to be investigated in a subsequent iteration. For each of these, in descending order of the adgroups budget, the bounding function for the subspace is calculated and compared to the current best solution and then branch on the node if necessary. The bound is calculated through using interior point method to solve the continuously relaxed keywords grouping model. If it can be established that the subspace cannot contain the optimal solution, the whole subspace is discarded, else it is checked whether the subspace consists of a better solution compared to the current best keywords grouping solution keeping the best of these. The search terminates when there is no unexplored parts of the solution space left, and the optimal solution is then the one recorded as "current best". For details about the solution space of branch and bound algorithm, see [13].

## 5. Experimental Validation

We conduct a set of computational experiments to validate our keywords grouping model and solution, based on two real-world datasets. The purposes of our experiments are two-fold. First, we aim to evaluate the effectiveness of our keywords grouping approach by comparing it with five baseline approaches with respect to profit, ROI[3] and the number of keywords assigned to adgroups. Second, we explore how the risk raised by our approach and the impact of advertiser's different risk-tolerances on keywords grouping decisions.

### 5.1 Data Descriptions

The first dataset (Dataset-1) is collected from advertising campaigns provided by an e-commerce firm who promoted celebration commodity on Amazon during the period from June 2016 to March 2017. This firm focuses on cake toppers and happy birthday decorations. In Dataset-1, we choose one of its major advertising campaigns, i.e., for Celebration Commodity Series, as our experimental subject. The ad campaign is titled "Celebration Commodity Series", and its setting is shown in Table A1 of Appendix. This campaign is divided into two adgroups by the advertiser originally, i.e., celebration banner (with keywords such as "happy birthday sign", "you are loved sign", "grand opening ribbon", "bunting banner", "pink happy birthday banner", etc.) and cake decoration (with keywords such as "black cake stand", "mickey cake topper", "wedding cake topper", "cake topper 1st birthday", "gold cake topper", etc.). The potential customers of the two adgroups have interests in celebration banners and cake decorating supplies, respectively. This ad campaign contains 90 keywords for two adgroups. The dataset contains records for keywords including keywords triggering the ad, the number of impressions, the number of clicks, CTR, CVR, average cost-per-click (CPC) and sales revenue. The mean and standard deviations of random factors (e.g., the keyword cost) can be estimated from historical reports and logs of advertising campaigns and the advertiser's proprietary information. Summary statistics for Dataset-1 are shown in Table 2.

The second dataset (Dataset-2) records advertising campaigns on Google AdWords by a large firm selling sportswear, from January 2016 to September 2016. In Dataset-2, the chosen campaign is titled as "Sneakers Series", and its setting is shown in Table A2 of Appendix. This ad campaign is divided into three adgroups by the advertiser originally, i.e., basketball (with keywords such as "basketball shoes", "cheap basketball shoes", "kids basketball shoes", "kobe basketball shoes", "high top basketball sneakers", etc.), running (with keywords such as "running shoes", "mens runners", "buy running shoes", "running sneakers", "running shoe online", "running shoes for men", etc.) and soccer (with keywords such as "soccer shoes", "indoor soccer shoes", "soccer cleats", "soccer boots", "kids soccer cleats", etc.). The potential customers of the three adgroups have interests in shoes for different types of sports. This dataset contains 305 keywords for three adgroups. Dataset-2 contains records for keywords identical to Dataset-1, the mean and standard deviations of random factors can be obtained in a similar way. Summary statistics for Dataset-2 are shown in Table 3.

---

[3] $ROI = E\left[\sum_{j=1}^{m}\sum_{i=1}^{n} x_{ij} d_i c_{ij}(r_{ij} v_i - p_{ij})\right] / E\left[\sum_{j=1}^{m}\sum_{i=1}^{n} x_{ij} d_i c_{ij} p_{ij}\right]$. The ROI is defined as the expected profit divided by the expected total cost.

The two datasets are quite rich to investigate the effectiveness of keywords grouping model and solution. We assume that there is no significant difference in ad quality for keyword-ad pairs, as this is a well-developed search advertising effort over multiple years.

### 5.2 Experimental Setup

The following experiments are set up as follows. For the first dataset, the total cost of these keywords in the chosen ad campaign is 19,200. In experiments on Dataset-1, we increase the total campaign budget from 2,000 to 20,000 by a step of 2,000, which is allocated to the two adgroups at the ratio of 2:1. For the second dataset, the total cost of these keywords in target ad campaign is 66,786. In experiments on Dataset-2, we increase the total campaign budget from 10,000 to 70,000 by a step of 10,000, which is allocated to the three adgoups at the ratio of 3:2:1. In the following experiments, the probability of chance constraint (i.e., $\alpha_j$) is set as 0.95. At different levels of campaign budget, the risk-tolerance (i.e., $\theta$) for risk-loving advertisers is $\theta = \infty$, and for risk-averse advertisers, $\theta = 0.3$.

### 5.3 Comparisons

We compare our approach (BBKG) with five baselines with respect to profit, ROI and the number of keywords assigned to adgroups. As far as we knew, there is limited research on keywords grouping and no comparative approach reported in the state-of-the-art literature. For comparison purposes, we implement two baseline approaches commonly used in practice and two baselines derived from the literature on keyword clustering, and the fifth is a deterministic approach derived from our approach. The first baseline represents the case that the advertiser puts all keywords into a single adgroup (i.e., BASE1-Nogrouping). The second subdivides the keywords according to products to be promoted by the advertiser (i.e., BASE2-Product). The third baseline approach (i.e., BASE3-Kcluster) is derived from a k-means clustering algorithm applied in [39] to understand the underlying intent of the query terms, which categorizes keywords with similar characteristics of onsite behaviors, such as pages per visit and click-through rate. In our context, the BASE3-Kcluster categorizes keywords with a set of characteristics associated with each referral keyword, including impressions, click-through rate, cost-per-click, conversion rate and value-per-sale. The fourth baseline approach (i.e., BASE4-Hierarchy) is derived from the keyword hierarchy [3]. Specifically, a domain-specific concept hierarchy is constructed on the basis of a high-quality Web directory such as Wikipedia, and then a keyword hierarchy is established by matching keywords with relevant concepts. Based on this keyword hierarchy, keywords can be grouped into several subsets related to different topics. The fifth baseline (i.e., BASE5-Profit) orderly assigns keywords into adgroups according to their profits obtained in a greedy manner following a deterministic model derived from our stochastic keywords grouping model developed in Section 4. In the following experiments, we assign keywords into adgroups using our solution proposed in Section 4 and five baselines independently. Note that our experiments are conducted based on the two realworld datasets about past advertising campaigns in laboratory.

Figure 1 show the profit and ROI obtained by our approach (BBKG) and five baselines at different levels of campaign budget on Dataset-1, respectively. Corresponding results on Dataset-2 are shown in Figure 2.

From Figures 1 and 2, we observe the following:

(1) On both Dataset-1 and Dataset-2, profits obtained by our approach and the five baselines increase with the total campaign budget. In general, with more budget available, more keywords are included to adgroups, and then more profit is generated.

(2) On both Dataset-1 and Dataset-2, our approach (BBKG) outperforms the five baselines in terms of the profit and ROI. This is because, on one hand, our approach can traverse more possibilities by considering uncertainties. On the other hand, there exist a few popular keywords that are of high profit but expensive. These baselines assign popular keywords to adgroups, instead of less-popular keywords with fair profit (or ROI). However, our approach based on the branch and bound algorithm can avoid such situation by traversing the solution space of keywords grouping decisions.

(3) On both Dataset-1 and Dataset-2, when the campaign budget is seriously limited, five baselines illustrate similar performance. This indicates that we can use the BASE1-Nogrouping to approximate other four baselines in the situation with a limited campaign budget; however, as the campaign budget increases, the other four baselines become superior to the BASE1-Nogrouping. In terms of ROI, BASE1-Nogrouping performs the worst. Undoubtedly, it suggests the necessity of keywords grouping decisions. As the campaign budget increases, more keywords are assigned. Under such complicated market environments, keywords grouping approaches show the strength to help advertisers obtain more profit.

(4) On Dataset-1, for our approach and the five baselines, with more campaign budget is allocated, while holding all others constant, will not certainly yield lower incremental per-unit returns, at some points; however, on Dataset-2, the marginal profit decreases with the campaign budget. In other words, the law of diminishing return in economics [48] does not necessarily work in keywords grouping problems. The possible explanation is that, high-profit keywords usually come with relatively high costs, which might be assigned as the campaign budget increases, rather than in low-budget situations. As such, the marginal profit of a keywords grouping solution may increase as the campaign budget increases in such situations. The difference in the marginal returns on the two datasets lies in the fact that experiments on Dataset-1 provide the limited campaign budget at the initial stage, while on Dataset-2, we initially allocate a relatively high campaign budget because the latter contains much more keywords than the former.

(5) On both Dataset-1 and Dataset-2, the ROI gap between our approach and five baselines are relatively large when the campaign budget is limited; while it declines as the campaign budget increases. This can be explained as follows. When the campaign budget becomes abundant, even if high-cost keywords are selected, there are still a lot of budgets left, so the negative influence on the profit due to the selection of high-cost keywords (by five baselines) is reduced. Thus, our approach and five baselines become closer as the campaign budget increases.

Figure 3 show the number of assigned keywords by our approach (BBKG) and five baselines on two datasets at different levels of campaign budget. From Figure 3, we observe the following phenomena.

(1) Overall, the number of assigned keywords increases with the campaign budget. At a given level of campaign budget, our approach (BBKG) assigns more keywords than the five baselines. The "long tail" phenomenon is of vital importance in search engine marketing [6]. Long-tail

keywords are referred as many less popular keywords employed by users to search the Internet. Our approach searches for the global optimum, which assigns many long-tail keywords into adgroups while skipping popular keywords.

(2) Similar to other four baselines, BASE5-Profit initially assigns fewer keywords than our approach, on both Dataset-1 and Dataset-2. However, the number of keywords assigned by BASE5-Profit increases more quickly than other baselines as the campaign budget increases, and gets close to our approach finally. This is because BASE5-Profit is a greedy strategy. It assigns the popular keywords into adgroups sequentially making the locally optimal choice. And high-profit keywords take up too much budget at the initial stage, which makes the BASE5-Profit only assign a limited number of keywords into adgroups when the campaign budget is very short. However, it solves the deterministic model derived from our approach. Thus, it can get similar assigned keyword number when the campaign budget is large enough.

From Figures 1-3, we can see that, on Dataset-1, although BASE2-Product assigns fewer keywords than BASE5-Profit, it obtains a similar profit; likewise, on Dataset-2, BASE3-Kcluster assigns fewer keywords than BASE2-Product, but it obtains a higher profit. We can learn that, for different approaches, it's not always the case that more keywords assigned in adgroups will certainly lead to higher profits.

**5.4 Risk-Tolerance**

Different keywords grouping approaches have different levels of risk under different levels of campaign budget. In this section we investigate the risks of our approach (BBKG) and five baselines at different levels of campaign budget. Experimental results show that it indicates that BBKG is superior to the five baselines, in that it can obtain the highest profit with relatively lower risk. Moreover, we also examine the effect of risk-tolerance in keywords grouping decisions by comparing our approach with BASE5-Profit. For each approach, we consider two versions of the approach with respect to different attitudes to risks: RiskAverse and RiskLoving. It's unsurprising to find that the higher risk-tolerance usually leads to more expected profits, which coincides with the rule of positive correlation between risk exposures and expected profits in financial economics [48]. In addition, our branch and bound algorithm for keywords grouping decisions can lead to an optimal solution. For details, refer to see Appendix A2.

# 6. Conclusions and Future Work

In this paper, we present a stochastic keywords grouping approach to maximize the expected profit, taking into account advertising budget constraints and risk-tolerance. Our model takes the CTR and CVR as random variables. Moreover, we developed a branch-and-bound algorithm solution for our keywords grouping model. Furthermore, experiments are conducted to validate the effectiveness of our model and solution, with two real-world datasets collected from reports and logs of historical search advertising campaigns. Experimental results showed that our keywords grouping approach outperforms five baselines with respect to the profit and ROI. As the campaign budget increases, our approach can approximately approach the optimum in a steady way, with relatively lower risks.

This research generates several interesting findings that illuminate critical managerial insights for advertisers in sponsored search. First, as the budget increases, our approach and the other four baselines become more superior to the BASE1-Nogrouping. This indicates that keywords grouping is a critical advertising decision that cannot be overlooked, especially under more complicated market environment with a large number of keywords.

Second, in keywords grouping decisions, as the budget increases, the profit grows accordingly; however, the marginal profit won't necessarily keep the decreasing trend. In other words, the marginal profit does not show the marginal diminishing effect, i.e., it does not always decrease with the increase of the budget. More specifically, in the case with a sufficient budget, the marginal profit usually decreases with the budget; while in the case with a limited budget, the marginal profit fluctuates with the budget. Thus, increasing the budget in keywords grouping decisions can be considered as a worthy try for advertisers to obtain additional profit.

Third, the optimal keywords grouping solution is essentially the outcome of a multifaceted trade-off. Advertisers need to comprehensively consider various factors (e.g., the budget, the advertiser's risk-tolerance, and keyword performance indexes) in order to get an optimal keywords grouping decision. Thus, assigning more keywords into adgroups or having more budget won't certainly lead to higher profits. This suggests a serious warning for advertisers that it's not wise to take the number of keywords as the criterion for keywords grouping decisions.

Lastly, as the budget increases, the risk fluctuates up and down continuously. Given that the budget is determined, the advertisers' attitude towards risk influences the optimal solution for

keywords grouping decisions. Our optimal keywords grouping approach can help advertisers gain more profit with relatively lower risks in sponsored search advertising.

In an ongoing work, we attempt to develop more sophisticated keywords grouping approaches, by considering a set of interwoven effects among keywords. Moreover, taking into account dynamics in search advertising markets, we plan to study keywords adjustments by tracking the realtime performance of advertising campaigns. Another interesting issue is to take into account other relevant factors (e.g., semantic relationship, region) in keywords decisions. Furthermore, we will explore the role of uncertainties in the keyword-related operations in sponsored search advertising.

# Acknowledgements

We are thankful to anonymous reviewers who provided valuable suggestions that led to a considerable improvement in the organization and presentation of this manuscript. This work was partially supported by NSFC (National Natural Science Foundation of China) grants (71672067, 71810107003).

# References

1. Abhishek, V. and Hosanagar, K. Keyword generation for search engine advertising using semantic similarity between terms. In *Proceedings of the ninth international conference on Electronic commerce*. Minneapolis, MN: ACM, 2007, pp. 89-94.
2. Abou Nabout, N. A novel approach for bidding on keywords in newly set-up search advertising campaigns. *European Journal of Marketing*, *49*, 5/6 (2015), 668-691.
3. Agarwal, A. and Mukhopadhyay, T. The impact of competing ads on click performance in sponsored search. *Information Systems Research*, 27, 3 (2016), 538-557.
4. Amaldoss, W.; Jerath, K.; and Sayedi, A. Keyword management costs and "broad match" in sponsored search advertising. *Marketing Science*, *35*, 2 (2015), 259-274.
5. Bing Ads. *Create a new ad group.* Available at https://help.bingads.microsoft.com/apex/index/3/en/53097. Accessed on May 23, 2019.

6. Brynjolfsson, E.; Hu, Y.; and Simester, D. Goodbye pareto principle, hello long tail: The effect of search costs on the concentration of product sales. *Management Science*, 57, 8 (2011), 1373-1386.

7. Burns, K. New Technology Briefing: Ten golden rules to search advertising. *Interactive Marketing*, *6*, 3 (2005), 248-252.

8. Chatwin, R. E. An overview of computational challenges in online advertising. In *2013 American Control Conference*. Washinton, USA: IEEE, 2013, pp. 5990-6007.

9. Chen, J. and Stallaert, J. An economic analysis of online advertising using behavioral targeting. *MIS Quarterly*, 38, 2 (2010), 429-449.

10. Chen, J.; Liu, D.; and Whinston, A. B. Auctioning keywords in online search. *Journal of Marketing*, *73*, 4 (2009), 125-141.

11. Chen, Y.; Xue, G. R.; and Yu, Y. Advertising keyword suggestion based on concept hierarchy. In *Proceedings of the 2008 international conference on web search and data mining*. Palo Alto, California: ACM, 2008, pp. 251-260.

12. Cholette, S.; Özlük, Ö.; and Parlar, M. Optimal keyword bids in search-based advertising with stochastic advertisement positions. *Journal of Optimization Theory and Applications*, *152*, 1 (2012), 225-244.

13. Clausen, J. Branch and bound algorithms-principles and examples. *Department of Computer Science, University of Copenhagen*, 1999, 1-30.

14. Du, X.; Su, M.; Zhang, X.; and Zheng, X. Bidding for Multiple Keywords in Sponsored Search Advertising: Keyword Categories and Match Types. *Information Systems Research*, *28*, 4 (2017), 711-722.

15. Edelman, B. and Ostrovsky, M. Strategic bidder behavior in sponsored search auctions. *Decision support systems*, 43, 1 (2007), 192-198.

16. Ghose, A. and Yang, S. An empirical analysis of search engine advertising: Sponsored search in electronic markets. *Management science*, 55, 10 (2009), 1605-1622.

17. Gong, J.; Abhishek, V.; and Li, B. Examining the Impact of Keyword Ambiguity on Search Advertising Performance: A Topic Model Approach. *MIS Quarterly*, 42, 3 (2018), 1-40.

18. Google AdWords. *How adgroups work.* Available at https://support.google.com/google-ads/answer/2375404?hl=en. Accessed on October 28, 2018.

*19.* Google AdWords. *Organize your account with ad groups.* Available at https://support.google.com/google-ads/answer/6372655?hl=en. Accessed on May 23, 2019.

20. Gopal, R.; Li, X.; and Sankaranarayanan, R. Online keyword based advertising: Impact of ad impressions on own-channel and cross-channel click-through rates. *Decision Support Systems*, 52, 1 (2011), 1-8.

21. Gotter, A. *How to Create the Most Effective Adgroups for Google AdWords.* Available at https://www.disruptiveadvertising.com/adwords/ad-groups/. Accessed on October 28, 2018.

22. Hillard, D.; Schroedl, S.; Manavoglu, E.; Raghavan, H.; and Leggetter, C. Improving ad relevance in sponsored search. In *Proceedings of the third ACM international conference on Web search and data mining.* New York: ACM, 2010, pp. 361-370.

23. Holthausen, D. M. and Assmus, G. Advertising budget allocation under uncertainty. *Management Science, 28*, 5 (1982), 487-499.

24. Hou, L. A hierarchical bayesian network-based approach to keyword auction. *IEEE Transactions on Engineering Management, 62*, 2 (2015), 217-225.

25. Interactive Advertising Bureau (IAB). *2018 IAB Internet Ad Revenue Full Year Report.* Available at https://www.iab.com/wp-content/uploads/2019/05/Full-Year-2018-IAB-Internet-Advertising-Revenue-Report.pdf. Accessed on May 10, 2019.

26. Jansen, B. J. and Clarke, B. Conversion potential: A metric for evaluating search engine advertising Performance. *Journal of Research in Interactive Marketing*, 11, 2 (2017), 142-159.

27. Jansen, B. J. *Understanding Sponsored Search: Core Elements of Keyword Advertising*. Cambridge: Cambridge University Press, 2011.

28. Jansen, B. J.; Sobel, K.; and Zhang, M. The brand effect of key phrases and advertisements in sponsored search. *International Journal of Electronic Commerce*, 16, 1 (2011), 77-106.

29. Kiritchenko, S. and Jiline, M. Keyword optimization in sponsored search via feature selection. In *New Challenges for Feature Selection in Data Mining and Knowledge Discovery*. Antwerp, Belgium: JMLR, 2008, pp. 122-134.

30. Kosuch, S. and Lisser, A. Upper bounds for the 0-1 stochastic knapsack problem and a B&B algorithm. *Annals of Operations Research,* 176, 1 (2010), 77-93.

31. Li, H.; Kannan, P. K.; Viswanathan, S.; and Pani, A. Attribution strategies and return on keyword investment in paid search advertising. *Marketing Science*, 35, 6 (2016), 831-848.

32. Lobo, M. S.; Vandenberghe, L.; Boyd, S.; and Lebret, H. Applications of second-order cone programming. *Linear algebra and its applications*, 284, 1-3 (1998), 193-228.

33. Lu, X. and Zhao, X. Differential effects of keyword selection in search engine advertising on direct and indirect sales. *Journal of Management Information Systems*, 30, 4 (2014), 299-326.

34. Luo, W.; Cook, D.; and Karson, E. J. Search advertising placement strategy: Exploring the efficacy of the conventional wisdom. *Information & Management,* 48, 8 (2011), 404-411.

35. Lutze, H. F. *The findability formula: The easy, non-technical approach to search engine marketing*. Hoboken, New Jersey: John Wiley & Sons, 2009.

36. Mohr, J. J.; Sengupta, S.; and Slater, S. F. *Marketing of High-technology Products and Innovations*. Upper Saddle River, NJ: Pearson Prentice Hall, 2010.

37. Muthukrishnan, S.; Pál, M.; and Svitkina, Z. Stochastic models for budget optimization in search-based advertising. *Algorithmica*, 58, 4 (2010), 1022-1044.

38. Nie, H.; Yang, Y.; and Zeng, D. Keyword Generation for Sponsored Search Advertising: Balancing Coverage and Relevance. *IEEE Intelligent Systems*, (2019), forthcoming.

39. Ortiz‐Cordova, A. and Jansen, B. J. Classifying web search queries to identify high revenue generating customers. *Journal of the American Society for Information Science and Technology*, *63*, 7 (2012), 1426-1441.

40. Pin, F. and Key, P. Stochastic variability in sponsored search auctions: observations and models. In *Proceedings of the 12th ACM conference on Electronic commerce.* San Jose, California: ACM, 2011, pp. 61-70.

41. Prekopa, A. *Stochastic Programming*. Dordrecht, Boston: Kluwer Academic Publishers, 1995.

42. Qiao, D.; Zhang, J.; Wei, Q.; and Chen, G. Finding competitive keywords from query logs to enhance search engine advertising. *Information & Management,* 54, 4 (2017), 531-543.

43. Ravi, S.; Broder, A.; Gabrilovich, E.; Josifovski, V.; Pandey, S.; and Pang, B. Automatic generation of bid phrases for online advertising. In *Proceedings of the third ACM international conference on Web search and data mining*. New York: ACM, 2010, pp. 341-350.

44. Regelson, M. and Fain, D. Predicting click-through rate using keyword clusters. In *Proceedings of the Second Workshop on Sponsored Search Auctions*. Ann Arbor, MI, Vol. 9623, 2006, pp. 1-6.

45. Rusmevichientong, P. and Williamson, D. P. An adaptive algorithm for selecting profitable keywords for search-based advertising services. In *Proceedings of the 7th ACM Conference on Electronic Commerce*. Ann Arbor, Michigan: ACM, 2006, pp. 260-269.

46. Rutz, O. J. and Bucklin, R. E. A model of individual keyword performance in paid search advertising. *Available at SSRN 1024765,* 2007.

47. Rutz, O. J.; Bucklin, R. E.; and Sonnier, G. P. A latent instrumental variables approach to modeling keyword conversion in paid search advertising. *Journal of Marketing Research*, 49, 3 (2012), 306-319.

48. Samuelson, P. A. and Nordhaus, W. D. *Microeconomics*. New York: McGraw-Hill/Irwin, 2001.

49. Sen, R. Optimal search engine marketing strategy. *International Journal of Electronic Commerce*, 10, 1 (2005), 9-25.

50. Shi, S. W. and Dong, X. The effects of bid pulsing on keyword performance in search engines. *International Journal of Electronic Commerce*, 19, 2 (2015), 3-38.

51. Shin, W. Keyword search advertising and limited budgets. *Marketing Science*, 34, 6 (2015), 882-896.

52. Skiera, B. and Abou Nabout, N. Practice prize paper—PROSAD: a bidding decision support system for profit optimizing search engine advertising. *Marketing Science*, 32, 2 (2013), 213-220.

53. Telang, R.; Rajan, U.; and Mukhopadhyay, T. The market structure for Internet search engines. *Journal of Management Information Systems*, 21, 2 (2004), 137-160.

54. Wächter, A.; Biegler, L. T. On the implementation of an interior-point filter line-search algorithm for large-scale nonlinear programming. *Mathematical programming,* 106, 1 (2006), 25-57.

55. Weber, I. *Top 3 Benefits of Using Keyword Grouping*. Available at https://seranking.com/blog/top-3-benefits-of-using-keyword-grouping/. Accessed on June 19, 2019.

56. Wiley, D. L. *Outsmarting Google: SEO Secrets to Winning New Business.* Pearson: Que Publishing, 2011.

57. WordStream. *AdWords Keywords grouping: How to Group Your Keywords in AdWords.* Available at https://www.wordstream.com/adwords-keyword-grouping. Accessed on October 28, 2018.

58. Wu, H.; Qiu, G.; He, X.; Shi, Y.; Shen, J.; Shen, J.; Bu, J.; and Chen, C. Advertising keyword generation using active learning. *International Conference on World Wide Web*. Madrid, Spain: ACM, 2009, pp. 1095-1096.

59. Xue, L.; Ray, G.; and Gu, B. Environmental uncertainty and IT infrastructure governance: A curvilinear relationship. *Information Systems Research*, 22, 2 (2011), 389-399.

60. Yang, Y.; Zhang, J.; Qin, R.; Li, J.; Liu, B.; and Liu, Z. Budget strategy in uncertain environments of search auctions: A preliminary investigation. *IEEE transactions on services Computing*, 6, 2 (2013), 168-176.

61. Yang, Y.; Qin, R.; Jansen, B. J.; Zhang, J.; and Zeng, D. Budget planning for coupled campaigns in sponsored search auctions. *International Journal of Electronic Commerce*, 18, 3 (2014), 39-66.

62. Yang, Y.; Zeng, D.; Yang, Y.; and Zhang, J. Optimal budget allocation across search advertising markets. *INFORMS Journal on Computing*, 27, 2 (2015), 285-300.

63. Yang, Y.; Yang, Y. C.; Jansen, B. J.; Lalmas, M. Computational Advertising: A Paradigm Shift for Advertising and Marketing?. *IEEE Intelligent Systems,* 32, 3 (2017), 3-6.

64. Yang, Y.; Li, X.; Zeng, D.; and Jansen, B. J. Aggregate effects of advertising decisions: a complex systems look at search engine advertising via an experimental study. *Internet Research*, 28, 4 (2018), 1079-1102.

65. Yang, Y.; Jansen, B. J.; Yang, Y., Guo, X.; and Zeng, D. Keyword optimization in sponsored search advertising: A multilevel computational framework. *IEEE Intelligent Systems*, 34, 1, (2019), 32-42.

66. Zenetti, G.; Bijmolt, T. H.; Leeflang, P. S.; and Klapper, D. Search engine advertising effectiveness in a multimedia campaign. *International Journal of Electronic Commerce*, 18, 3 (2014), 7-38.

67. Zhang, X. and Feng, J. Cyclical bid adjustments in search-engine advertising. *Management Science*, *57*, 9 (2011), 1703-1719.

68. Zhang, Y.; Zhang, W.; Gao, B.; Yuan, X.; and Liu, T. Y. Bid keyword suggestion in sponsored search based on competitiveness and relevance. *Information Processing & Management*, 50, 4 (2014), 508-523.

69. Zhou, Y. and Naroditskiy, V. Algorithm for stochastic multiple-choice knapsack problem and application to keywords bidding. In *Proceedings of the 17th international conference on World Wide Web.* Beijing: ACM, 2008, pp. 1175-1176.

70. Zhou, Y.; Huang, F.; and Chen, H. Combining probability models and web mining models: a framework for proper name transliteration. *Information Technology and Management,* 9, 2 (2008), 91-103.

## Table 1. List of Notations

| Notation | Definition |
|---|---|
| $d_i$ | the search demand of the $i^{th}$ keyword |
| $c_{ij}$ | the click-through rate (CTR) of the $i^{th}$ keyword in the $j^{th}$ adgroup |
| $r_{ij}$ | the conversion rate (CVR) of the $i^{th}$ keyword in the $j^{th}$ adgroup |
| $v_i$ | the value-per-sale (VPC) of the $i^{th}$ keyword |
| $p_{ij}$ | the cost-per-click (CPC) of the $i^{th}$ keyword in the $j^{th}$ adgroup |
| $B_j$ | the budget constraint of the $j^{th}$ adgroup |
| $x_{ij}$ | the decision variable indicated whether the $i^{th}$ keyword is assigned to the $j^{th}$ adgroup |
| $n$ | the number of keywords |
| $m$ | the number of adgroups |
| $s_i$ | the cost of the $i^{th}$ keyword |
| $\alpha_j$ | the prescribed probability of budget constraint for the $j^{th}$ adgroup |
| $\varphi^{-1}(\alpha_j)$ | the Inverse Gaussian distribution density function of $\alpha_j$ |
| $\mu_i$ | the mean of $s_i$ |
| $\sigma_i$ | the standard deviation of $s_i$ |
| $\theta$ | the risk-tolerance of an advertiser |

## Table 2. Summary Statistics of Dataset-1

| Variable | Search demand ($d_i$) | CTR ($c_{ij}$) | CVR ($r_{ij}$) | VPC ($v_i$) | CPC ($p_{ij}$) | Keyword cost($s_i$) |
|---|---|---|---|---|---|---|
| Mean | 1211.90 | 0.04 | 0.53 | 16.31 | 0.30 | 2.13 |
| SD | 2296.07 | 0.15 | 0.37 | 14.64 | 0.08 | 3.67 |

## Table 3. Summary Statistics of Dataset-2

| Variable | Search demand ($d_i$) | CTR ($c_{ij}$) | CVR ($r_{ij}$) | VPC ($v_i$) | CPC ($p_{ij}$) | Keyword cost($s_i$) |
|---|---|---|---|---|---|---|
| Mean | 289.57 | 0.17 | 0.35 | 21.90 | 1.15 | 8.95 |
| SD | 2279.9 | 0.23 | 0.57 | 54.53 | 0.4 | 41.22 |

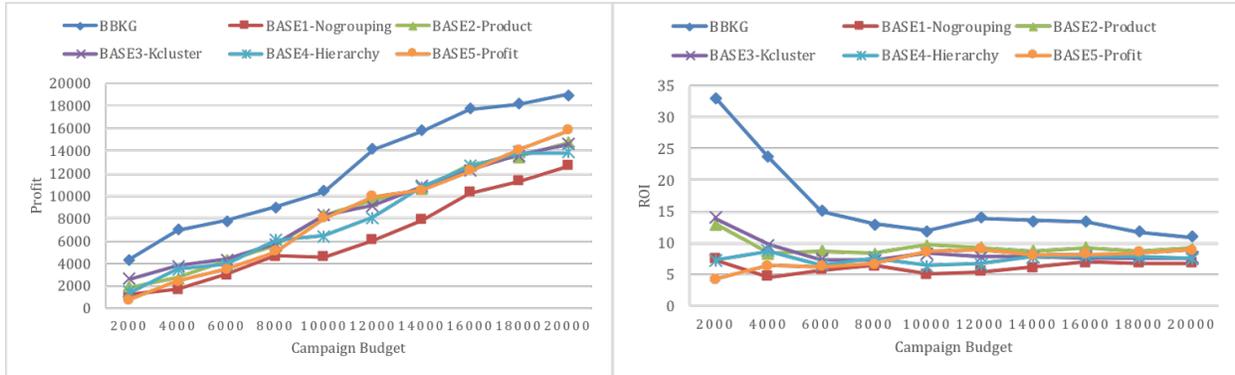

**Figure 1. Profit and ROI Obtained by BBKG and Five Baselines on Dataset-1**

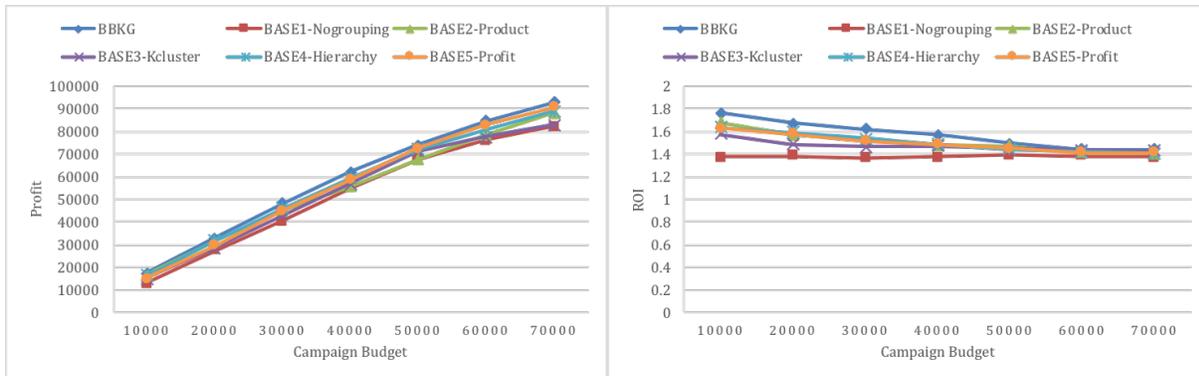

**Figure 2. Profit and ROI Obtained by BBKG and Five Baselines on Dataset-2**

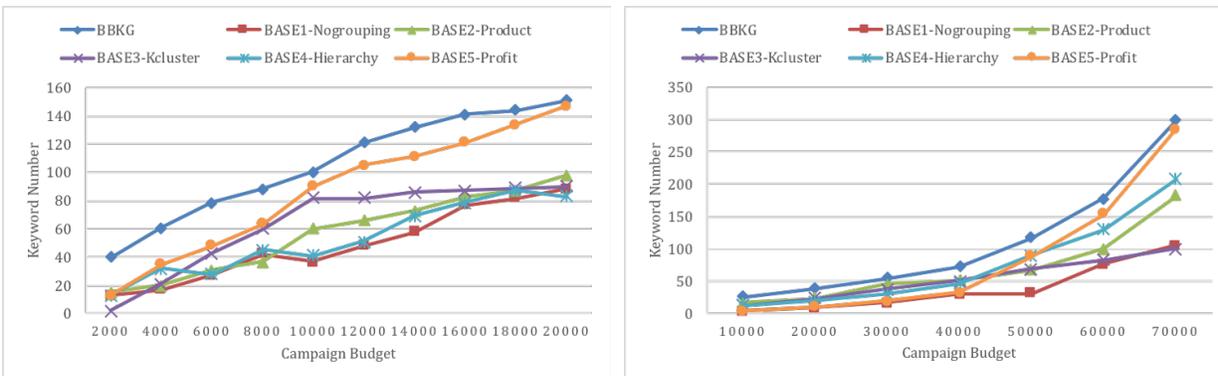

(a) Dataset-1　　　　　　　　　　　　(b) Dataset-2

**Figure 3. The Number of Keywords Assigned by BBKG and Five Baselines**

# Appendix for "Optimal Keywords Grouping in Sponsored Search Advertising under Uncertain Environments"

### A.1. Campaign Setting

**Table A1. Campaign Setting of Dataset-1 (on Amazon Advertising)**

| Setting | Choice | Description |
| --- | --- | --- |
| Campaign name | Celebration Commodity Series | The campaign name that the advertiser chooses will be seen in Campaign Manager. |
| Targeting type | automatic targeting | Targeting uses keywords and products to show advertiser's ads on search and detail pages to relevant shoppers. For Sponsored Products campaigns, advertiser can create two types of targeting: automatic and manual. |
| Bidding strategy | dynamic bids - up and down | Search advertising platform increases advertiser's bids in realtime for keywords that may be more likely to convert to a sale, and reduces bids for keywords that are less likely to convert to a sale. |

Source: Amazon Advertising. *Create a Sponsored Products campaign*. Available at https://advertising.amazon.com/help#GJUCNANNV3GQVXJZ. Accessed on June 21, 2019.

**Table A2. Campaign Setting of Dataset-2 (on Google AdWords)**

| Setting | Choice | Description |
| --- | --- | --- |
| Campaign name | Sports Shoes Series | A name for advertiser's campaign. |
| Campaign type | search network campaign | It tailors the campaign setup to show ads on Google.com to get more visitors to advertiser's website. |
| Networks | include search partners | It indicates that advertiser want her ad to appear in a network of sites having special contracts with Google, to display Google search ads on their search results. |
| Delivery method | standard | It evenly distributes the budget over time. |
| Dd scheduling | all day | The start and end date of the campaign |
| Bidding | automated bid strategy | It is offered by Google AdWords focusing on getting conversions |

Source: Google AdWords. *About campaign settings*. Available at https://support.google.com/google-ads/answer/1704395?co=ADWORDS.IsAWNCustomer%3Dfalse&hl=en. Accessed on June 21, 2019.

## A.2. Risk-Tolerance

Figure A1 shows the risks of our approach (BBKG) and five baselines at different levels of campaign budget on the two datasets, respectively.

From Figure A1, we can observe that, on Dataset-1, our approach initially (i.e., in the case with the campaign budget as 20) shows a pretty high risk that is nearly same as BASE2-Product, but it quickly decreases and becomes lower than five baselines; while on Dataset-2, compared to five baselines, in most cases, our approach shows a relatively low risk. Combining with Figures 1 and 2, it indicates that BBKG is superior to the five baselines, in that it can obtain the highest profit with relatively lower risk.

Among five baselines, BASE5-Profit is the only one that considers the advertiser's risk-tolerance. As a matter of fact, BASE5-Profit uses a deterministic model transformed from the stochastic keywords grouping model (i.e., Equation 1) proposed in this research. Thus, we examine the effect of risk-tolerance in keywords grouping decisions by comparing our approach with BASE5-Profit. For each approach, we consider two versions of the approach with respect to different attitudes to risks: RiskAverse and RiskLoving. The former is referred to the optimal solution of model (1), while the latter is a special case of the former with $\theta = \infty$.

Figure A2 shows the profit and corresponding risk by our approach (BBKG), for risk-loving (with risk-tolerance $\theta = \infty$) and risk-averse advertisers (with risk-tolerance $\theta=0.3$), at different levels of campaign budget, on the two datasets, respectively.

Figure A3 shows the profit and corresponding risk by our approach (BBKG) and BASE5-Profit, for risk-averse advertisers (with risk-tolerance $\theta=0.3$) at different levels of campaign budget, on the two datasets, respectively.

From Figures A2 and A3, we observe the following:

(1) For a risk-loving advertiser, the Risk-Loving strategy apparently performs better than the Risk-Averse strategy in terms of the expected profit. The reason behind is that, the risk for the Risk-Averse strategy is below the advertiser's risk-tolerance, which constraints the expected profit. The risk attitude plays an important role in keywords grouping decisions. Specifically, risk-tolerances influence the determination of the optimal solutions in keywords grouping decisions, and the higher risk-tolerance usually leads to more expected profits. It coincides with the rule of positive correlation between risk exposures and expected profits in financial economics.

(2) For a risk-averse advertiser, both our keywords grouping approach and the BASE5-Profit obtain acceptable performance. Moreover, our approach is relatively superior (i.e. the higher profit). It implies that our branch and bound algorithm for keywords grouping decisions can lead to an optimal solution, compared with the greedy algorithm employed in BASE5-Profit.

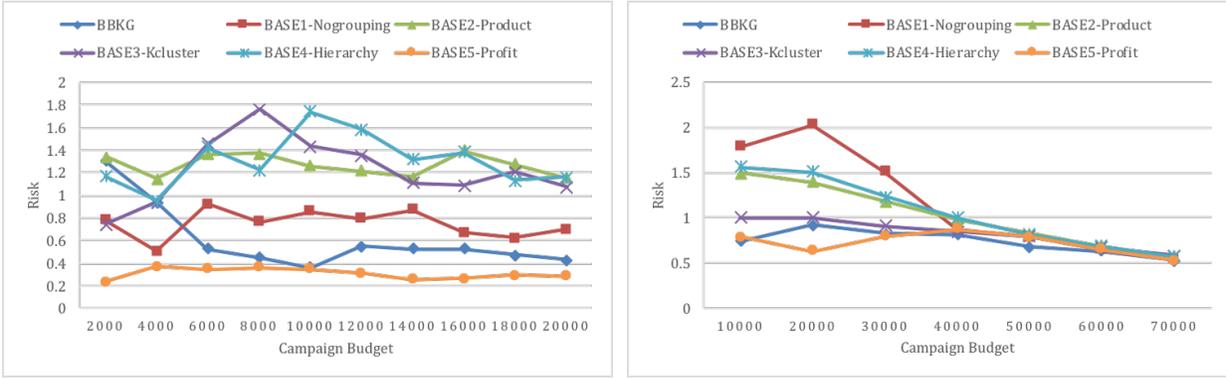

(a) Dataset-1                              (b) Dataset-2

**Figure A1. Risk over Campaign Budget by BBKG and Five Baselines**

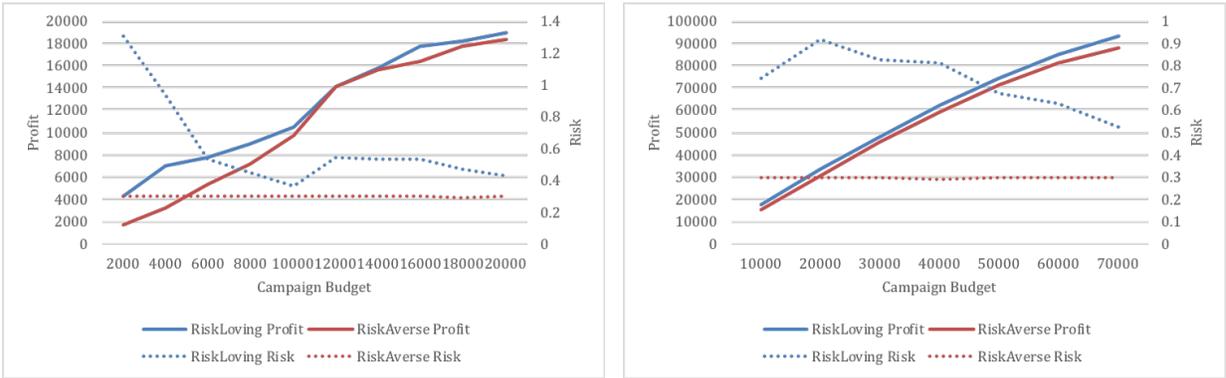

(a) Dataset-1                              (b) Dataset-2

**Figure A2. Profit and Risk for Risk-Loving ($\theta = \infty$) and Risk-Averse Advertisers ($\theta = 0.3$) over Campaign Budget by BBKG**

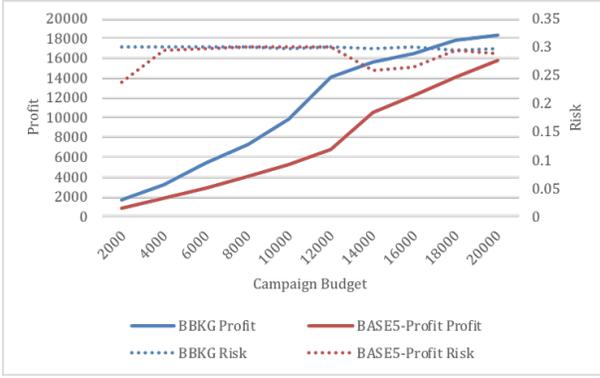 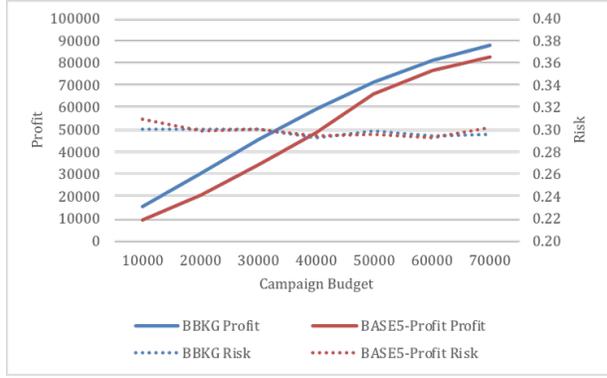

(a) Dataset-1      (b) Dataset-2

**Figure A3. Profit and Risk for Risk-Averse Advertisers ($\theta = 0.3$) over Campaign Budget by BBKG and BASE5-Profit**